\documentclass{article}

\usepackage{microtype}
\usepackage{graphicx}
\usepackage{subfigure}
\usepackage{booktabs} 

\usepackage{hyperref}

\usepackage[accepted]{MOSS_camera_ready}

\usepackage{amsmath}
\usepackage{amssymb}
\usepackage{mathtools}
\usepackage{amsthm}

\usepackage[capitalize,noabbrev]{cleveref}

\theoremstyle{plain}

\theoremstyle{definition}

\theoremstyle{remark}

\usepackage[textsize=tiny]{todonotes}

\icmltitlerunning{Evaluating Generalization and Representation Stability}

\begin{document}

\onecolumn
\icmltitle{Evaluating Generalization and Representation Stability in Small LMs via Prompting, Fine-Tuning and Out-of-Distribution Prompts}



\icmlsetsymbol{equal}{*}

\begin{icmlauthorlist}
\icmlauthor{Rahul Raja}{cmu,linkedin,stanford}
\icmlauthor{Arpita Vats}{bu,linkedin}
\end{icmlauthorlist}

\icmlaffiliation{linkedin}{LinkedIn, California, USA}
\icmlaffiliation{bu}{Boston University, Boston, USA}
\icmlaffiliation{cmu}{Carnegie Mellon University, Pittsburgh, USA}
\icmlaffiliation{stanford}{Stanford University, Palo Alto, USA}

\icmlcorrespondingauthor{Rahul Raja}{rauhl.110392@gmail.com}
\icmlcorrespondingauthor{Arpita Vats}{arpita.vats09@gmail.com}

\icmlkeywords{Machine Learning, ICML}

\vskip 0.3in


\printAffiliationsAndNotice{Work does not relate to position at Linkedin.} 

\begin{abstract}
We investigate the generalization capabilities of small language models under two popular adaptation paradigms: few-shot prompting and supervised fine-tuning. While prompting is often favored for its parameter efficiency and flexibility, it remains unclear how robust this approach is in low-resource settings and under distributional shifts. This paper presents a comparative study of prompting and fine-tuning across task formats, prompt styles, and model scales, with a focus on their behavior in both in-distribution and out-of-distribution (OOD) settings.

Beyond accuracy, we analyze the internal representations learned by each approach to assess the stability and abstraction of task-specific features. Our findings highlight critical differences in how small models internalize and generalize knowledge under different adaptation strategies. This work offers practical guidance for model selection in low-data regimes and contributes empirical insight into the ongoing debate over prompting versus fine-tuning. Code for the experiments is available at the following link.\footnote{\url{https://anonymous.4open.science/r/moss-small-lm-experiments-BD03/moss_small_lm.ipynb}}

\end{abstract}

\section{Introduction}
\label{submission}

Few-shot prompting and supervised fine-tuning are two widely adopted strategies for adapting pretrained language models (LMs) to downstream tasks. Prompting adapts models by conditioning on in-context examples at inference time without updating model parameters \cite{brown2020language}, whereas fine-tuning involves directly optimizing the model on labeled data. While prompting is attractive for its flexibility and efficiency, its reliability in low-resource settings—particularly for small-scale language models like GPT-2 \cite{radford2019language} and DistilGPT2 \cite{sanh2019distilbert}—remains uncertain.

In this work, we present a systematic comparison of prompting and fine-tuning using three GPT-2 variants: \texttt{distilgpt2}, \texttt{gpt2}, and \texttt{gpt2-medium}, evaluated across a suite of language understanding tasks. We investigate three central questions: (1) How does prompting performance scale with the number of in-context examples compared to fine-tuning under an equivalent data budget? (2) How well does each method generalize to out-of-distribution (OOD) prompt templates? (3) How stable are their internal representations across prompt variations?

We begin with prompting on synthetic multi-task tasks (sentiment, grammar correction, arithmetic, plural forms), followed by fine-tuning on IMDb sentiment classification. In addition to standard accuracy comparisons, we use t-SNE to analyze the structure of prompt and hidden-layer representations across models and prompt styles.

This study reveals important distinctions in how small LMs internalize supervision under different adaptation strategies, with implications for generalization and representational robustness in low-data regimes.

\section{Scaling and Shot Count: Accuracy and Representation Geometry}

To understand how model scale affects few-shot generalization, we evaluated three GPT-2 variants—\texttt{distilgpt2}, \texttt{gpt2}, and \texttt{gpt2-medium}—on synthetic classification tasks with varying numbers of in-context examples. Each model was prompted with $k \in \{1, 2, 3, 4, 5\}$ labeled examples and queried on a held-out test instance. The predicted label $\hat{y}$ was derived from the model's completion of the prompt
\[
P = [x_1, y_1; \ldots; x_k, y_k; x_{\text{test}},\ \_?],
\]
where the goal is to infer the label $y_{\text{test}}$. As expected, accuracy improved with increasing $k$, with \texttt{gpt2-medium} consistently outperforming its smaller counterparts.

However, a t-SNE visualization ~\cite{vanDerMaaten2008tsne} of the final hidden state embeddings from each model (see Appendix~\ref{appendix:tsne-prompt}, Figure~\ref{fig:tsne-prompt-model}) reveals an interesting asymmetry: although \texttt{gpt2-medium} achieved the highest task classification accuracy, its prompt representations did not form the most visually distinct clusters in the 2D projection. In contrast, \texttt{gpt2} (base) exhibited clearer task-level separation under t-SNE, suggesting that the embeddings of larger models may become more abstract or compressed, making them less amenable to low-dimensional geometric partitioning.

This decoupling between performance and representational separability implies that scaling improves behavioral accuracy but does not necessarily enhance interpretability via dimensionality-reduced visualization. These findings highlight that few-shot generalization involves both behavioral alignment and latent structuring—two aspects that do not always co-occur in small-scale diagnostic settings.

\begin{figure}[ht]
\vskip 0.1in
\centering
\subfigure[Accuracy across shot counts.]{
    \includegraphics[width=0.45\textwidth]{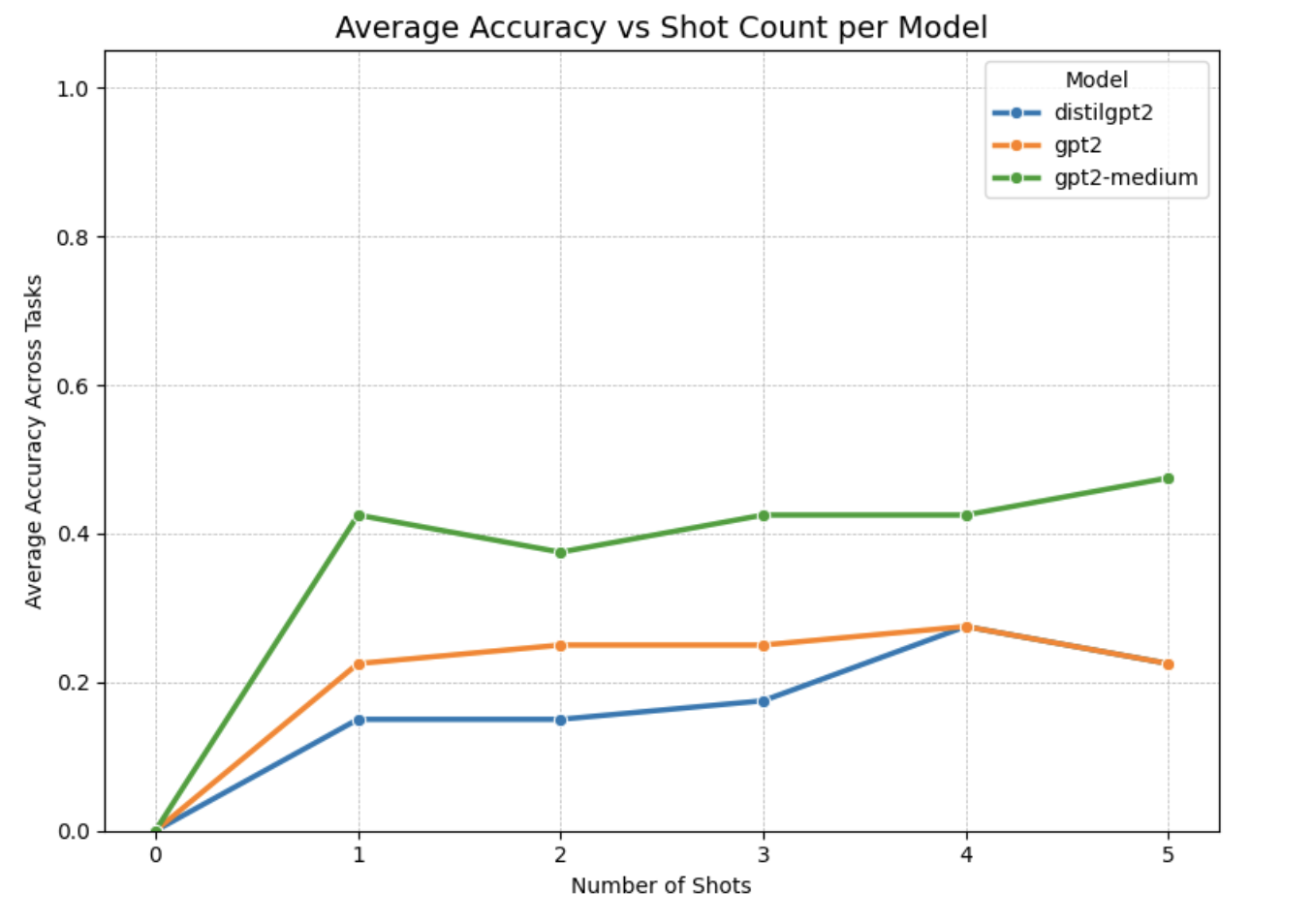}
    \label{fig:accuracy-shot}
}
\hfill
\subfigure[Prompting vs fine-tuning.]{
    \includegraphics[width=0.45\textwidth]{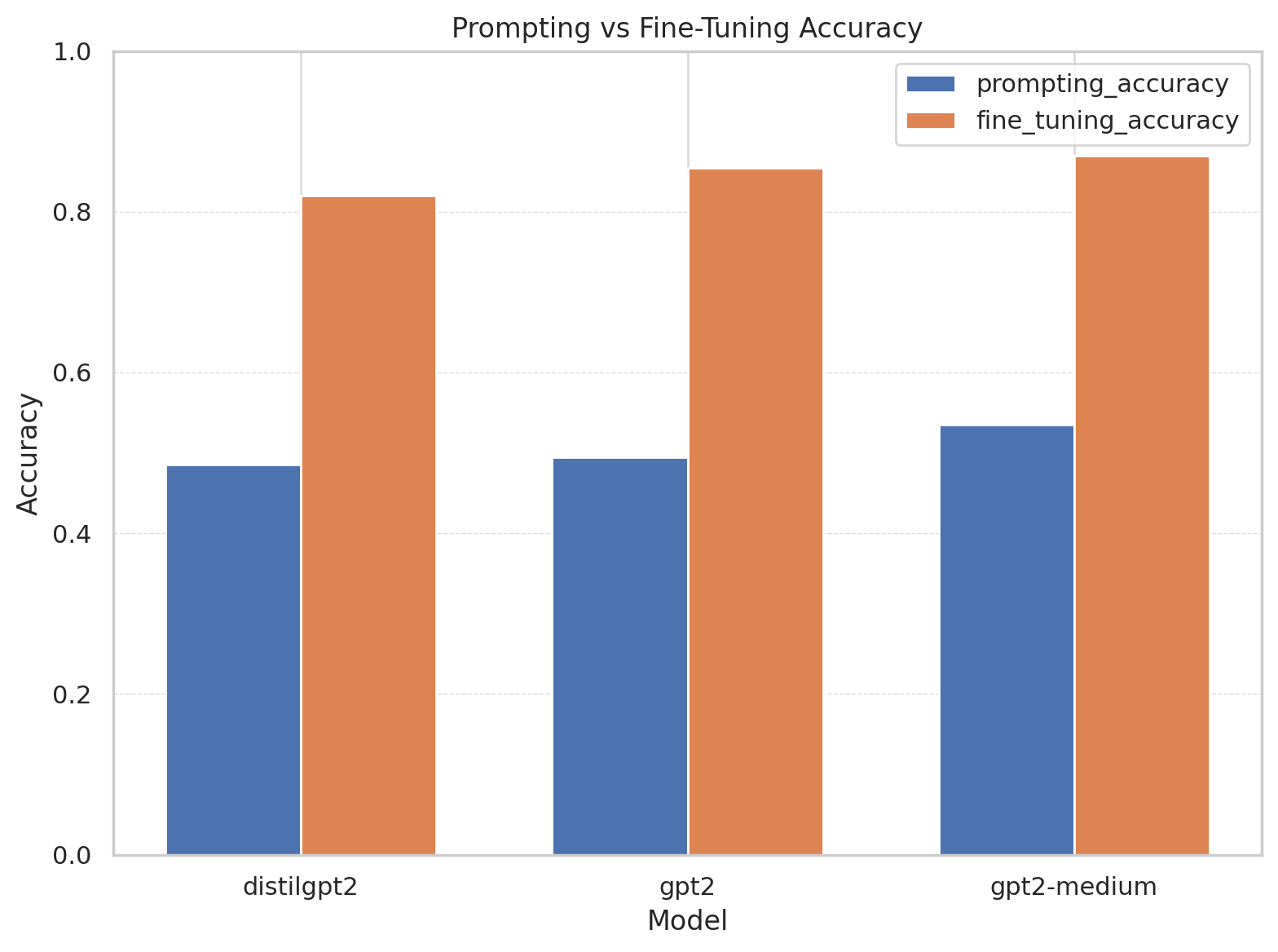}
    \label{fig:prompt-vs-fine}
}
\caption{Comparison of (a) few-shot accuracy trends and (b) training strategies under the same data budget.}
\label{fig:combined-comparison}
\vskip -0.2in
\end{figure}

\section{Prompting vs. Fine-Tuning Accuracy}

To quantify the performance gap between prompting and parameter tuning, we evaluate three GPT-2 variants on a binary sentiment classification task using the IMDb dataset. A random sample of 1000 examples is drawn and split into training and test subsets (\(80\%/20\%\)). Each model is exposed to a fixed budget of \(k=5\) labeled examples from the training set, used under two distinct paradigms.

In the first, \textbf{prompting}, the examples \( \{(x_i, y_i)\}_{i=1}^5 \) are formatted as in-context demonstrations within a prompt. The model completes the sequence for a test query \( x_{\text{test}} \) to produce a label prediction \( \hat{y} \), without modifying model parameters. In the second, \textbf{fine-tuning}, the same examples are used to optimize the cross-entropy loss via parameter updates:
\[
\mathcal{L}_{\text{FT}} = -\frac{1}{k} \sum_{i=1}^{k} \log p_\theta(y_i \mid x_i),
\]
where \( \theta \) denotes the model weights.

The table below summarizes accuracy scores on the held-out test set:

\begin{center}
\begin{tabular}{lcc}
\toprule
\textbf{Model} & \textbf{Prompting Accuracy} & \textbf{Fine-Tuning Accuracy} \\
\midrule
\texttt{distilgpt2} & 0.485 & 0.820 \\
\texttt{gpt2}       & 0.495 & 0.855 \\
\texttt{gpt2-medium}& 0.535 & 0.870 \\
\bottomrule
\end{tabular}
\end{center}

Fine-tuning consistently outperforms prompting across all scales, with gains exceeding 30 absolute percentage points. This highlights the efficacy of even minimal gradient-based adaptation when data is scarce. Prompting accuracy improves modestly with model size, reflecting an increase in pre-trained capacity for in-context reasoning. However, these gains remain small relative to the improvements achieved via fine-tuning. The performance gap, defined as \( \Delta = \mathcal{A}_{\text{FT}} - \mathcal{A}_{\text{Prompt}} \), is approximately constant across models and underscores that parameter tuning contributes orthogonal benefits to scaling.

These findings also carry implications for lightweight models. The smallest variant, despite having limited in-context learning ability, achieves a substantial accuracy increase through fine-tuning. This suggests that even low-capacity models retain latent flexibility when supervised adaptation is allowed. In practical scenarios such as edge deployment or privacy-sensitive fine-tuning, small models may still deliver competitive performance when gradient access is available.

In summary, while prompting offers convenience, fine-tuning remains the dominant approach in small-data regimes particularly when performance is critical and modest training is feasible.

\section{Generalization to Out-of-Distribution (OOD) Prompts}

We investigate how model behavior changes under variations in prompt formulation, using three semantically aligned but structurally distinct formats: \textit{standard}, \textit{QA-style}, and \textit{good/bad}. While the underlying task—binary sentiment classification on IMDb—remains fixed, these prompt types differ in phrasing and label tokens \cite{reynolds2021prompt}. Prompt construction details are provided in Appendix~\ref{appendix:ood-setup}.

Let \( x \in \mathcal{X} \), \( y \in \{0, 1\} \), and \( \pi_p \) be the prompt formatter for type \( p \in \mathcal{P} \). For in-context prompting, a \( k \)-shot input is constructed as:
\[
\texttt{Prompt} = \big\|_{i=1}^k \pi_p(x_i, y_i) \, \big\| \, \pi_p(x_{\text{test}},\ \text{?}),
\]
where ``\text{?}'' indicates a missing label to be predicted by the model. The operator \( \| \) denotes newline-concatenation. Model output is parsed to extract predicted tokens, mapping ``positive''/``good'' to 1 and ``negative''/``bad'' to 0.

\begin{figure}[ht]
\vskip 0.2in
\centering
\subfigure[Prompting: sensitivity to label phrasing]{
    \includegraphics[width=0.45\linewidth]{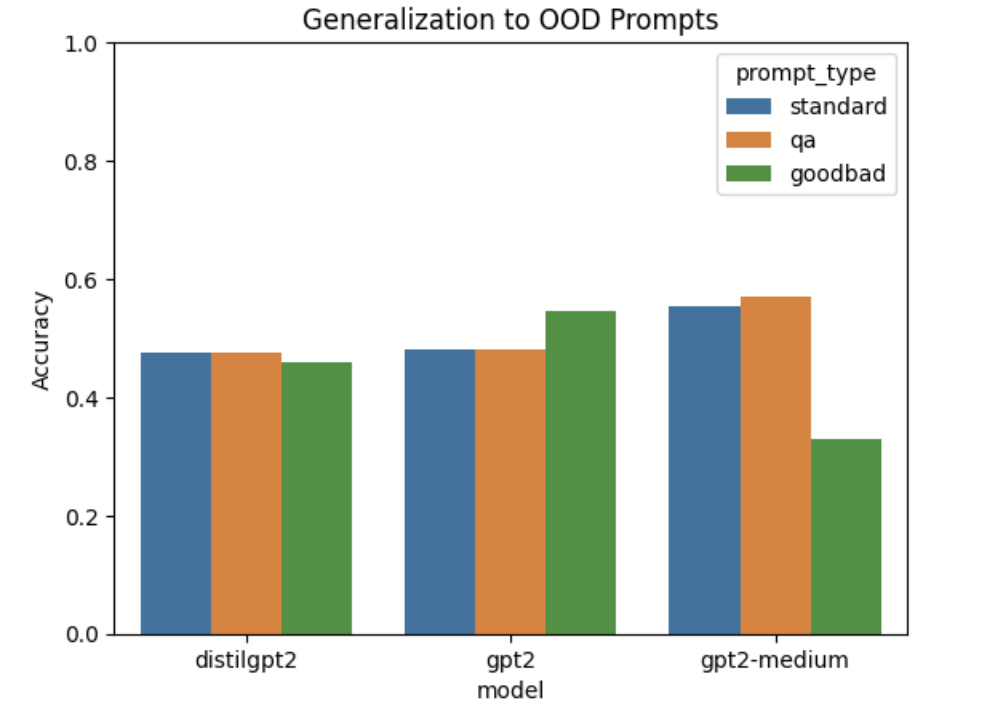}
    \label{fig:ood-prompting}
}
\hfill
\subfigure[Fine-tuning: robustness across formats]{
    \includegraphics[width=0.45\linewidth]{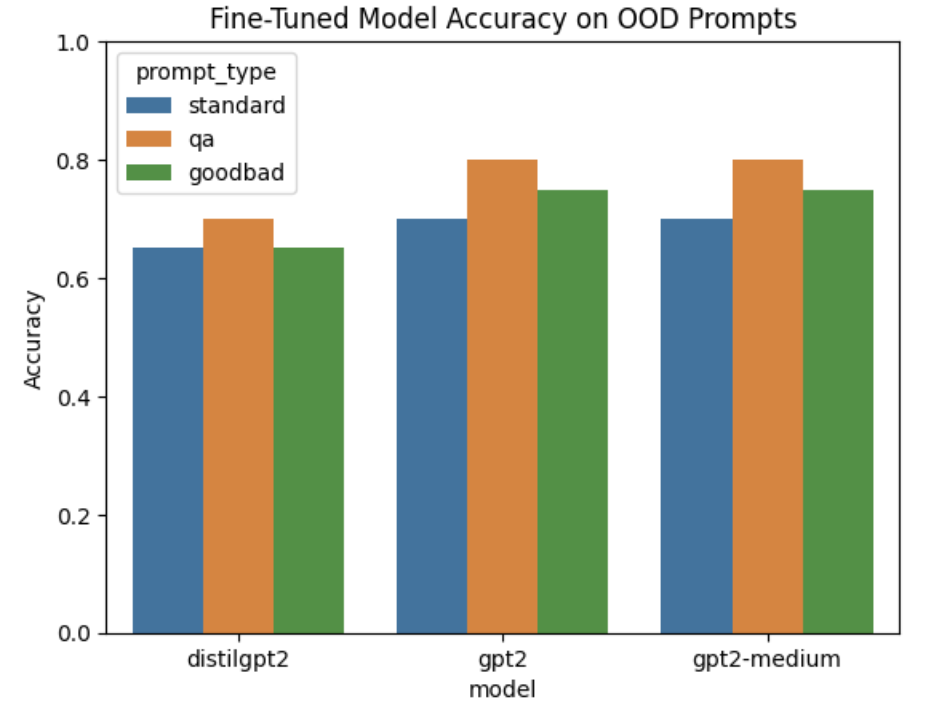}
    \label{fig:ood-finetuned}
}
\caption{Accuracy across prompt formats using (a) in-context prompting and (b) fine-tuned classifiers. Fine-tuning improves robustness to surface variation.}
\label{fig:ood-comparison}
\vskip -0.2in
\end{figure}

Prompting results, shown in Figure~\ref{fig:ood-comparison}(a), reveal that all models maintain consistent accuracy between the \textit{standard} and \textit{QA-style} formats. However, performance drops sharply for the \textit{good/bad} format, especially in \texttt{gpt2-medium}, suggesting strong dependence on lexical alignment \cite{webson2022prompt}. Interestingly, \texttt{distilgpt2} is more stable, possibly due to reduced reliance on pretraining-specific surface cues.

We next test whether fine-tuning improves robustness to such prompt variations. Each model is fine-tuned using the standard prompt, then evaluated across all prompt types. Given a prompt \( \pi_p(x) \), prediction is computed as:
\[
\hat{y} = \arg\max f_\theta(\texttt{Tokenizer}(\pi_p(x))),
\]
where \( f_\theta \) is the fine-tuned model. As shown in Figure~\ref{fig:ood-comparison}(b), accuracy remains high across all prompt variants. The performance gap seen in prompting vanishes, indicating that fine-tuning enables models to internalize task semantics beyond surface form, improving generalization to OOD prompts.

\section{Probing Internal Representations with Linear Classifiers}

We investigate the extent to which different GPT-2 model variants encode task-discriminative information in their internal representations. Using a set of synthetic prompts from three task types—sentiment analysis, arithmetic addition, and factual QA—we extract final-layer hidden states from each model and evaluate whether a simple linear classifier can recover task identity.

Given an input sequence \( x \), we obtain its representation \( \mathbf{h} \in \mathbb{R}^d \) by mean-pooling the final-layer hidden states:
\[
\mathbf{h} = \frac{1}{T} \sum_{t=1}^T \mathbf{h}_t^{(L)},
\]
where \( \mathbf{h}_t^{(L)} \) is the hidden state of token \( t \) from the final layer and \( T \) is the number of tokens in the input. These representations are used to train a logistic regression classifier to predict the task label.

To assess stability, we run the experiment 10 times with different train-test splits and report the mean classification accuracy. We observe a consistent improvement with model scale: \texttt{distilgpt2} achieves an average accuracy of 67\%, \texttt{gpt2} reaches 60\%, and \texttt{gpt2-medium} peaks at 73\%.

These results suggest that internal representations become increasingly task-aware and linearly decodable with scale. To better understand the geometric structure of these embeddings, we visualize them using t-SNE in the Appendix (Figure~\ref{fig:tsne-combined-appendix}). The plots confirm that larger models form more compact and separable clusters for each task category, providing qualitative support for the quantitative probe accuracy results.

\section{Layer-wise Probing of Task Representations}

To understand how task-discriminative information emerges across the model depth, we perform layer-wise probing on each GPT-2 variant. For each transformer layer, we extract the mean-pooled hidden state from synthetic task inputs and train a logistic regression classifier to predict the task category (sentiment, arithmetic, QA). This allows us to evaluate the linear decodability of task identity at each layer.

\begin{figure}[ht]
\vskip 0.2in
\centering
\includegraphics[width=0.8\linewidth]{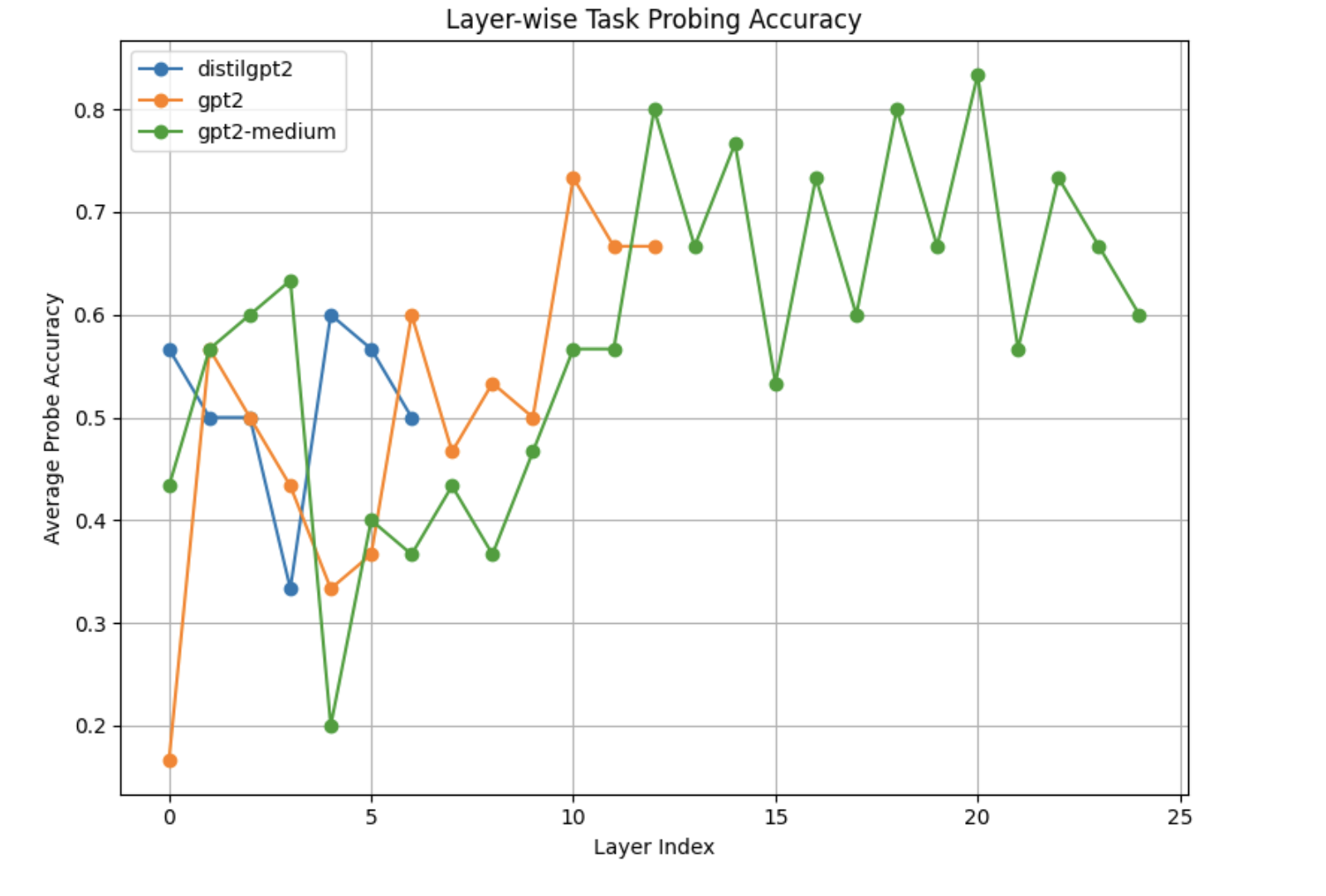}
\caption{Layer-wise probe accuracy for each GPT-2 variant. Accuracy improves monotonically with depth in larger models, while \texttt{distilgpt2} peaks earlier.}
\label{fig:layer-accuracy}
\vskip -0.1in
\end{figure}

Figure~\ref{fig:layer-accuracy} shows the average probe accuracy per layer, aggregated over 10 runs. We observe that deeper layers consistently encode more linearly separable task information. For \texttt{gpt2-medium}, accuracy climbs sharply after the midpoint and saturates in the final layers. In contrast, \texttt{distilgpt2} achieves peak probe accuracy in the middle layers, suggesting a compressed representational depth.

These results support the view that representational abstraction accumulates with depth and scale, and that later layers consolidate semantic structure in a form more accessible to simple classifiers.

\section{Conclusion and Future Directions}

This study demonstrates that small-scale diagnostic experiments can reveal nuanced trends in language model representations and generalization. We show that task identity becomes more linearly decodable with scale and depth, yet visualization via t-SNE reveals that larger models do not always exhibit cleaner geometric separation—highlighting a disconnect between abstraction and interpretability. Our probing results confirm that later layers encode more task-relevant information, and fine-tuning yields greater robustness to prompt variation than in-context learning alone.

Future work may explore richer probing techniques (e.g., CKA, RSA), test robustness under syntactic or cross-lingual shifts, and investigate whether modular or retrieval-augmented models can improve both interpretability and performance at smaller scales. These directions support the case for using small, controlled setups to build principled understanding of model behavior.

\nocite{langley00}

\bibliography{example_paper}
\bibliographystyle{icml2025}

\section{Appendix}

\subsection{Prompt Construction and OOD Evaluation Details}
\label{appendix:ood-setup}

To evaluate generalization to prompt distribution shifts, we conducted few-shot prompting experiments across three distinct prompt styles: \textit{standard}, \textit{QA-style}, and \textit{good/bad}. All experiments used the IMDb dataset, from which a random sample of 1000 reviews was selected. The data was split into 80\% training and 20\% test sets using stratified sampling to preserve label balance.

\subsubsection*{Labeling Across Prompt Types}

Each review in the IMDb dataset is annotated with a binary label:  
\[
y \in \{0, 1\}, \quad \text{where } 1 = \text{positive}, \; 0 = \text{negative}.
\]

Depending on the prompt style, the label is rendered differently:

\begin{itemize}
    \item In \textbf{standard} and \textbf{QA-style} prompts, label 1 is represented as \texttt{positive} and 0 as \texttt{negative}.
    \item In the \textbf{good/bad} prompt variant, label 1 is mapped to \texttt{good} and label 0 to \texttt{bad}.
\end{itemize}

This variation in surface form allows us to probe how sensitive models are to label-space shifts, despite task semantics remaining constant.

\subsubsection*{Few-Shot Prompt Format}

Each model is prompted with \(k = 4\) labeled examples from the training set, formatted according to the chosen prompt style. These demonstrations are concatenated to a test input to form the final prompt:

\[
P = \texttt{[demo}_1\texttt{]} \, \Vert \cdots \Vert \, \texttt{[demo}_k\texttt{]} \, \Vert \, \texttt{[test\_input]},
\]

where \( \Vert \) denotes newline-separated concatenation. The prompt is passed to the model using a text-generation pipeline, and the generated output is parsed to extract the final predicted label token.

\subsubsection*{Prompt Style Examples}

Below are concrete examples for the same test instance under each style:

\begin{itemize}
    \item \textbf{Standard Prompt:}
\begin{verbatim}
Review: The movie was surprisingly emotional and beautifully directed.
Sentiment:
\end{verbatim}

    \item \textbf{QA-style Prompt:}
\begin{verbatim}
Question: What is the sentiment of this review?
"The movie was surprisingly emotional and beautifully directed."
Answer:
\end{verbatim}

    \item \textbf{Good/Bad Prompt:}
\begin{verbatim}
Review: The movie was surprisingly emotional and beautifully directed.
Label:
\end{verbatim}
\end{itemize}

\subsubsection*{Prediction and Evaluation}

The model’s output is parsed to extract the last match of a keyword (``label'', ``sentiment'', or ``answer'') followed by a candidate class token. Predicted tokens are normalized and interpreted as:

\[
\texttt{positive/good} \rightarrow 1, \quad \texttt{negative/bad} \rightarrow 0.
\]

These predictions are then compared to ground-truth test labels to compute accuracy. This design isolates prompt-induced performance variation while holding data and task objective constant.

\subsection{t-SNE of Prompt Representations (Per Model)}
\label{appendix:tsne-prompt}

To further investigate the structure of model representations across different prompt types, we conduct a t-SNE analysis using four distinct tasks: \textit{sentiment}, \textit{addition}, \textit{grammar correction}, and \textit{plurality classification}. For each model—\texttt{distilgpt2}, \texttt{gpt2}, and \texttt{gpt2-medium}—we extract the final-layer mean-pooled hidden states for prompts belonging to these tasks and visualize them using t-SNE (perplexity = 5).

\begin{figure}[ht]
\vskip 0.2in
\centering
\includegraphics[width=0.95\linewidth]{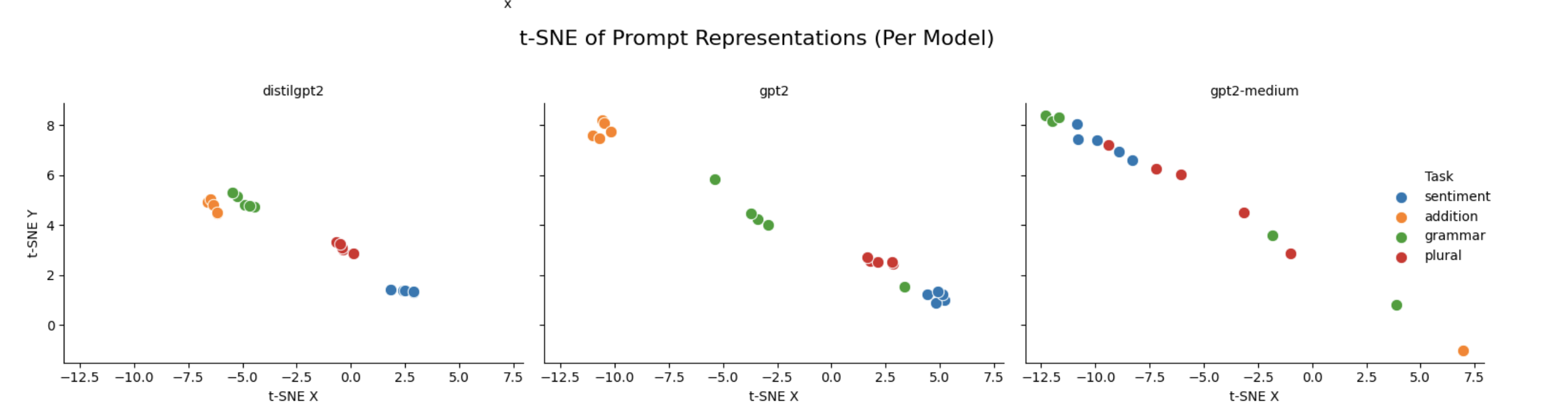}
\caption{t-SNE visualization of prompt representations across four tasks: sentiment (blue), addition (orange), grammar (green), and plural (red). Each column corresponds to a different GPT-2 model.}
\label{fig:tsne-prompt-model}
\vskip -0.1in
\end{figure}

\textbf{Observations:} The visualization highlights the evolution of representational geometry with model scale:

\begin{itemize}
  \item \textbf{distilgpt2:} Clusters are generally compact and well-separated, but task boundaries are rigid and uniform. This suggests some hard-coded task awareness, possibly due to compression and distilled pretraining. However, the inter-cluster margins are modest.
  
  \item \textbf{gpt2:} Task clusters begin to spread out, reflecting more nuanced and semantically differentiated representations. However, some overlap persists between \textit{grammar} and \textit{plural}, which may indicate shared syntactic cues in those prompts.
  
  \item \textbf{gpt2-medium:} The largest model exhibits smooth task transitions and distributed separability—task types are more linearly disentangled, with broader cluster geometry. Notably, unlike smaller models, \texttt{gpt2-medium} doesn't over-cluster by prompt type, instead forming gradients of semantic similarity.
\end{itemize}

These trends suggest a shift from discrete partitioning in smaller models to more expressive and abstract task encoding in larger ones. Moreover, the behavior of \texttt{gpt2-medium} implies that as scale increases, the model builds richer internal manifolds where prompt intent is represented more smoothly and generalizably across related tasks.

Overall, this analysis reinforces earlier findings: larger models do not just memorize templates—they develop latent structures that organize prompts semantically, which may explain their stronger generalization to unseen prompt types or task variants.

\subsection{t-SNE Visualization of Internal Representations}

To understand how internal representations encode task-specific structure, we visualize the mean-pooled hidden states from the final transformer layer of each model using t-SNE. For each synthetic input belonging to one of three task categories—sentiment classification, arithmetic addition, or question answering—we extract its representation and project it to two dimensions using t-SNE (perplexity = 3).

\begin{figure}[ht]
\vskip 0.2in
\centering
\includegraphics[width=0.95\linewidth]{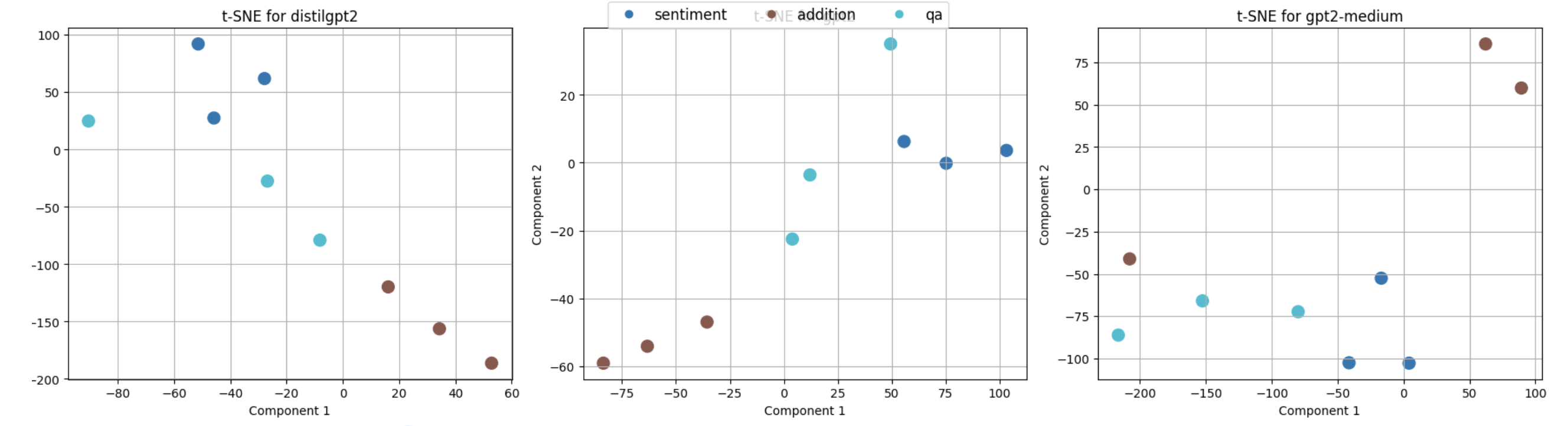}
\caption{t-SNE visualization of final-layer mean-pooled hidden representations for \texttt{distilgpt2}, \texttt{gpt2}, and \texttt{gpt2-medium}. Each point represents a task input and is colored by task category.}
\label{fig:tsne-combined-appendix}
\vskip -0.1in
\end{figure}

\textbf{Insights:} The visualization reveals a clear progression in representational structure with increasing model size. In \texttt{distilgpt2}, the points show partial overlap between classes, indicating that representations are somewhat entangled. For \texttt{gpt2}, the clusters begin to separate, but still exhibit intermixing at the boundaries. By contrast, \texttt{gpt2-medium} displays well-separated clusters with tighter intra-task grouping and clearer inter-task margins.

This qualitative trend aligns with the quantitative results in the probe accuracy table: the more distinct the clusters, the higher the performance of a linear classifier trained to distinguish tasks. These findings suggest that larger models not only encode richer semantics but also organize them in a more geometrically separable fashion—making them more accessible to lightweight downstream classifiers.

Moreover, the fact that even \texttt{distilgpt2} produces partially separable clusters supports the hypothesis that transformer models begin to form task-aware abstractions early in scale, which become increasingly linearly decodable as capacity increases.

\subsection{Appendix: Layer-wise Task Probing}

To complement our analysis of final-layer representations, we evaluate probe accuracy at every individual layer of each model. For each synthetic input, we extract mean-pooled hidden states from all transformer layers. At each layer index \( l \), a logistic regression classifier is trained to predict the task label using the corresponding representation \( \mathbf{h}^{(l)} \). Accuracy is averaged over 10 randomized train-test splits.

\textbf{Insights:} In \texttt{distilgpt2}, task-relevant features are most decodable around layers 5–6, after which accuracy flattens or declines—possibly due to representational compression. For \texttt{gpt2}, accuracy rises steadily with depth, while \texttt{gpt2-medium} shows a dramatic increase in the second half of the network, with peak accuracy approaching 90\% in the final layers.

These trends suggest that deeper and larger models progressively refine and separate task structure as information flows forward. Additionally, the sharp gains in \texttt{gpt2-medium} highlight the benefit of scale in both capacity and depth for learning linearly decodable abstractions.


\end{document}